\documentclass[acmtog]{acmart}

\copyrightyear{2024}
\acmYear{2024}
\setcopyright{rightsretained}
\acmConference[SA Posters '24]{SIGGRAPH Asia 2024 Posters}{December 3--6,
2024}{Tokyo, Japan}
\acmBooktitle{SIGGRAPH Asia 2024 Posters (SA Posters '24), December 3--6,
2024, Tokyo, Japan}\acmDOI{10.1145/3681756.3698208}
\acmISBN{979-8-4007-1138-1/24/12}

\acmSubmissionID{141}

\usepackage{graphicx}

\usepackage{booktabs}

\usepackage{multirow}
\usepackage{amsmath,bm}
\usepackage{url}
\usepackage[ruled]{algorithm2e} %

\SetAlFnt{\small}
\SetAlCapFnt{\small}
\SetAlCapNameFnt{\small}
\SetAlCapHSkip{0pt}
\usepackage{subcaption}
\usepackage{float}

\author{Pascal Clausen}
\authornote{Authors contributed equally to this work.}
\email{pascal.clausen@scanlinevfx.com}
\affiliation{
    \institution{Netflix Eyeline Studios}
    \country{Canada}
}

\author{Li Ma}
\authornotemark[1]
\email{lmaag@connect.ust.hk}
\affiliation{
    \institution{Netflix Eyeline Studios}
    \country{USA}
}

\author{Mingming He}
\authornotemark[1]
\email{hmm.lillian@gmail.com}
\affiliation{
    \institution{Netflix Eyeline Studios}
    \country{USA}
}

\author{Ahmet Levent Taşel}
\authornotemark[1]
\email{leventtasel@gmail.com}
\affiliation{
    \institution{Netflix Eyeline Studios}
    \country{Canada}
}

\author{Oliver Pilarski}
\authornotemark[1]
\authornote{Corresponding author}
\email{oliver.pilarski@scanlinevfx.com}
\affiliation{
    \institution{Netflix Eyeline Studios}
    \country{Germany}
}

\author{Paul Debevec}
\email{debevec@gmail.com}
\affiliation{
    \institution{Netflix Eyeline Studios}
    \country{USA}
}

\begin{document}

\title{Fitting Spherical Gaussians to Dynamic HDRI Sequences}

\begin{abstract}
We present a technique for fitting high dynamic range illumination (HDRI) sequences using anisotropic spherical Gaussians (ASGs) while preserving temporal consistency in the compressed HDRI maps. Our approach begins with an optimization network that iteratively minimizes a composite loss function, which includes both reconstruction and diffuse losses. This allows us to represent all-frequency signals with a small number of ASGs, optimizing their directions, sharpness, and intensity simultaneously for an individual HDRI. To extend this optimization into the temporal domain, we introduce a temporal consistency loss, ensuring a consistent approximation across the entire HDRI sequence.
\end{abstract}

\maketitle

\keywords{Dynamic HDRI fitting, evironment lighting, anisotropic spherical Gaussians}

\begin{figure}[t]
    \centering
    \includegraphics[width=\linewidth]{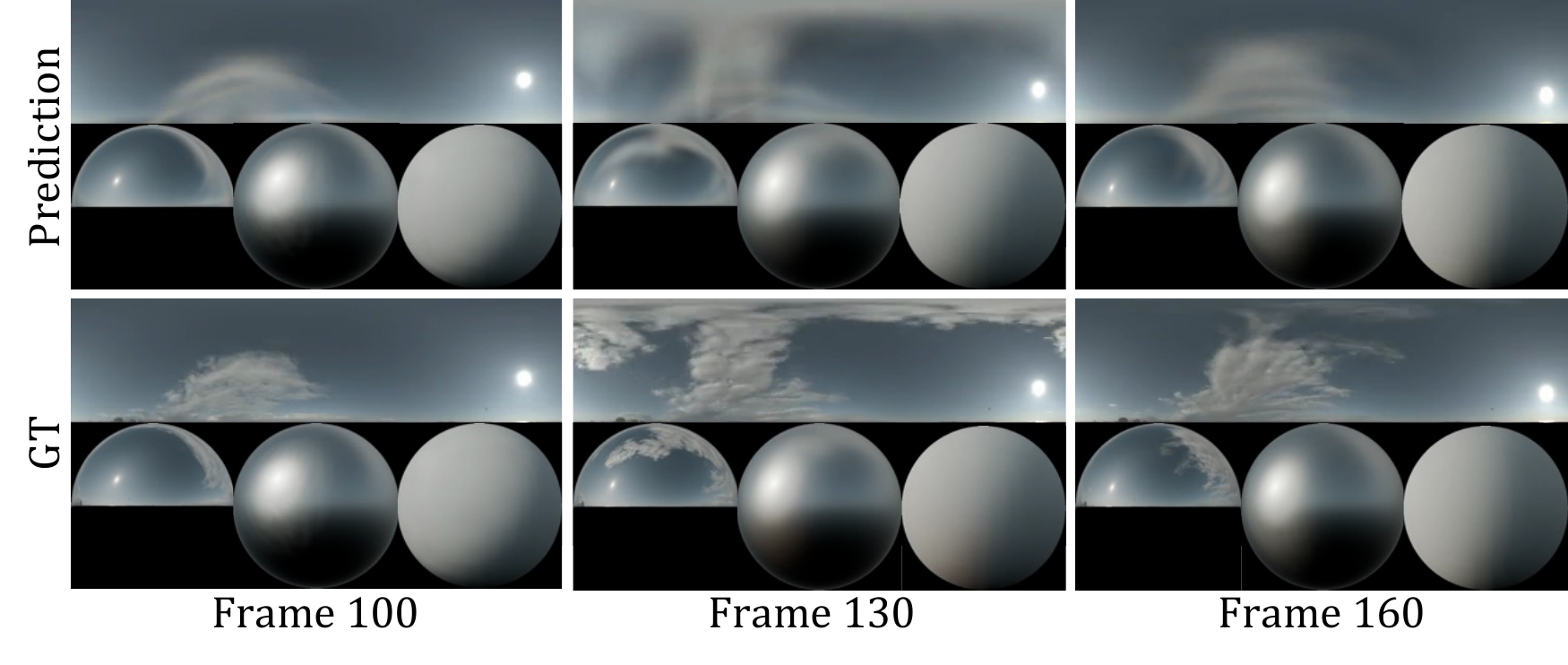}
    \caption{Our method approximates an HDRI sequence using a temporally consistent mixture of anisotropic spherical Gaussians (ASGs). Below are three example frames showing our fitting results and the rendering of balls with varying roughness under the corresponding HDRI.}
    \label{fig:teaser}
\end{figure}

\section{Introduction}

Environment lighting is commonly used to simulate distant illumination from a 360-degree scene. In the context of compressed lighting representations, ~\citet{Ramamoorthi2001_sh} use spherical harmonics (SH) for a linear approximation of low-frequency lighting. While this method is compact and efficient, it is limited to low-frequency lighting. To address higher frequencies, \citet{Ng2003_wavelets} introduce a nonlinear wavelet basis that retains only the largest terms, representing lighting across all frequencies. However, this method struggles with changing viewpoints and handles only diffuse surface interactions. \citet{tsai2003_sg} propose using spherical radial basis functions (SRBFs) to model environments, capturing high-frequency signals through nonlinear optimization with scattered SRBFs.

Spherical Gaussians (SGs), like SH and SRBF, are rotation-invariant and compact, capable of representing both sharp and diffuse lights with relatively few parameters, enabling high compression. \citet{xu_asg_2013} emphasize the user-friendly control of anisotropic SG (ASG) parameters and their efficient precomputation of light transfer functions. Yet, we observe, that L2 loss based optimizations (\cite{tsai2003_sg, xu_asg_2013}) performs poorly with low-frequency lighting. We propose a new loss function that combines L1 reconstruction loss with a diffuse loss to better preserve all frequencies. While the rotation-invariance of ASGs provides stable approximations for rotated HDRIs, frame-by-frame approximation of HDRI sequences, such as in (\cite{Stumpfel2004_sunandsky}), lacks temporal stability. To address this issue, we introduce a  temporal consistency term, ensuring stable and consistent HDRI sequences over time.

In summary, this work introduces a technique to approximate a time sequence of high dynamic range natural illumination using a small number of ASGs, with the following technical contributions:

\begin{enumerate}
    \item An optimization algorithm featuring a novel loss function that combines diffuse loss and L1 reconstruction loss, effectively capturing both low-frequency and high-frequency information in HDRIs.
    \item An extension of the optimization to the temporal domain, enabling the approximation of temporally consistent environment lighting.
\end{enumerate}

\section{Design and Implementation}
To approximate an HDRI using a mixture of ASGs, we represent each frame of the HDRI sequence with a set of ASGs, keeping the number of ASGs constant but varying their parameters. For each ASG, we use the same definition as the \textit{Bingham distribution} in \cite{xu_asg_2013}. Each ASG $G_i(\mathbf{d})$ is defined by the formula:
\begin{equation}
G_i( \mathbf{d}; [\mu, \lambda], [\mathbf{u}, \mathbf{v}], \mathbf{c} ) = \mathbf{c}  e^{ - \mu (\mathbf{d} \cdot \mathbf{u}) - \lambda (\mathbf{d} \cdot \mathbf{v}) } \text{, }
\end{equation}
where $\mathbf{d}$ is the query direction vector, $\mathbf{u}, \mathbf{v}$ being the tangent and bi-tangent axes, respectively. $[\mu, \lambda]$ are the bandwidths along the two axes, indicating the sharpnesses in these directions. $\mathbf{c}$ denotes the peak intensity of the ASG. We parameterize $[\mathbf{u}, \mathbf{v}]$ by $[\mathbf{u}, \mathbf{n}]$, where 
$\mathbf{n}$ if the direction of the gaussian lobe and $\mathbf{v}$ is derived from those directions. Each frame of the HDRI sequence is represented by a mixture of ASGs: $I_{pred}(\mathbf{d}) = \sum_{i}{G_i(\mathbf{d})}$, where $I_{pred}$ denotes the predicted HDRI. To optimize temporally consistent ASGs, our algorithm involves the following steps:

\begin{itemize}
\item For the initial frame of the HDRI time sequence, we optimize the ASG parameters (directions, sharpness, intensity) using an Adam optimizer for 24'000 epochs.
\item For each consecutive frame, we initialize the ASG parameters with those from the previous frame and then optimize them using an Adam optimizer with an additional temporal regularization loss for 6'000 epochs.
\end{itemize}

At each optimization iteration, we densely sample $256\times512$ pixels from the HDRI with equirectangular projection, and optimize the Gaussian parameters $\mathbf{g}_i = \left(\mu_i, \lambda_i, \mathbf{u}_i, \mathbf{n}_i, \mathbf{c}_i\right)$ using an analysis-by-synthesis approach. We minimize a composite loss function $L=w(\alpha L_R + \beta L_D) + \gamma L_T$, where $w$ is the solid angle of each pixel. We empirically set $\alpha=1.0$, $\beta=1.0$ and $\gamma=0.5$ in our experiments. $L_R$, $L_D$, $L_T$ are the reconstruction loss, diffuse loss, and temporal consistency loss given by:
\setlength{\belowdisplayskip}{3pt} 
\setlength{\abovedisplayskip}{3pt} 
\begin{align*}
L_R &= \|I_{pred} - I_{gt}\|_1\text{, }
L_D = \|D_{pred} - D_{gt}\|_1, \\
L_T &= {\textstyle \sum_i} \|(\mathbf{g}^{t}_i - \mathbf{g}^{t-1}_i) / \max_i\{\mathbf{g}^{t-1}_i\}\|_2 \text{,}
\end{align*}

where $I_{pred}$ and $I_{gt}$ are the predicted and ground truth HDRI intensities at the sampled directions. Our experiments show that using L1 loss effectively fits sharp edges. $D_{pred}$ and $D_{gt}$ denote the predicted and ground truth intensities of a diffused HDRI version at the sampled directions, which is a Lambertian light map approximated using spherical harmonics (SH) of degree 3, with SH coefficients obtained through numerical integration. The diffuse loss is crucial for optimizing ASGs to maintain correct total energy. $\mathbf{g}^{t}_i$ and $\mathbf{g}^{t-1}_i$ are parameters of $i$-th ASG for the current and previous frame.

\section{Results}
\begin{figure}[!htb]
  \centering
    \includegraphics[width=0.9\linewidth]{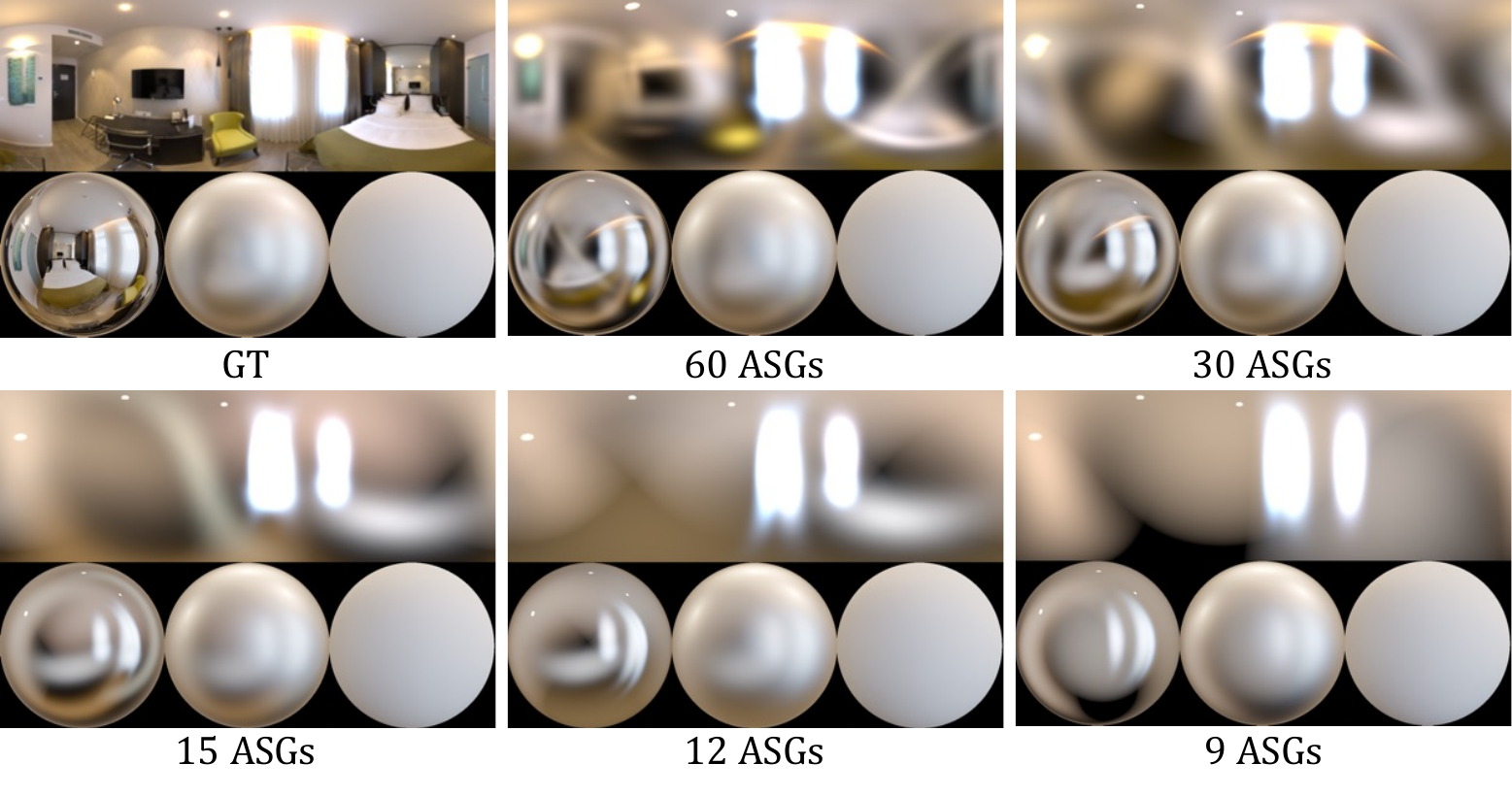}
    \caption{The comparison of the ground truth HDRI and compression results with different numbers of ASGs, shown through latlongs and renderings.
    }
    \label{fig:dynamic_numsg}
\end{figure}

We compare the fitting results of HDRIs with their ground truth using different numbers of ASGs, shown as ball renderings in Fig.~\ref{fig:dynamic_numsg}. It can be observed that beyond a small number of ASGs (e.g. 15), the ground truth differences in the rendered balls at most levels of roughness become negligible, when using the compressed HDRI for environment lighting. This indicates that our method effectively reconstructs the HDRI with a relatively low number of ASGs. Moreover, the fidelity of the rendering results across different roughness levels indicates that the compressed HDRI is well preserved for different frequencies. Our method is also applicable to fitting with isotropic spherical Gaussians. Please refer to the accompanying video for more results.

\begin{figure}[!htb]
  \centering
    \includegraphics[width=\linewidth]{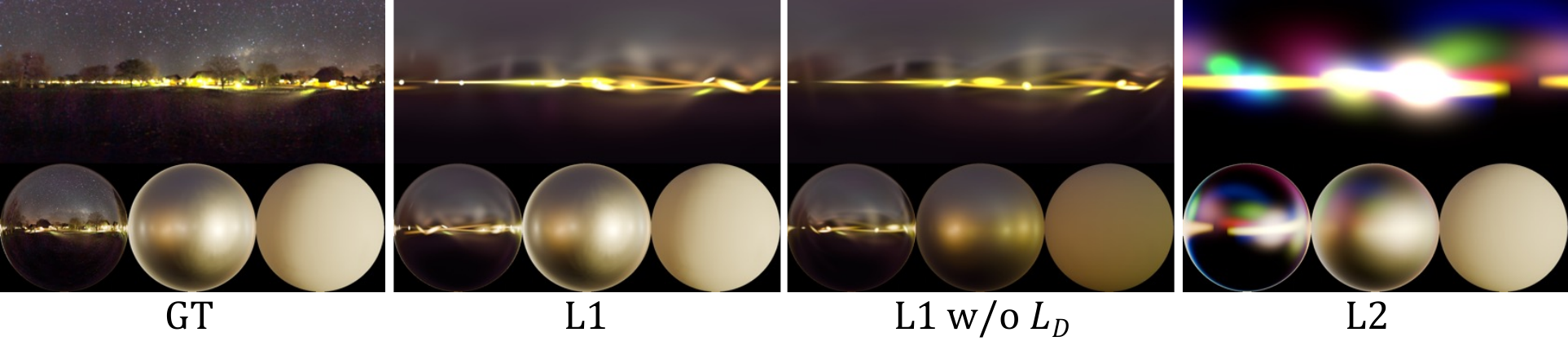}
    \caption{The comparison of different loss functions used for HDRI fitting with 15 ASGs. L1 refers to the L1 loss function used in the reconstruction loss; L2 indicates that the L1 loss is replaced by L2 loss; $L_D$ is the diffuse loss.}
  \label{fig:dynamic_lossfunc}
\end{figure}

\begin{figure}[!htb]
    \centering
    \includegraphics[width=\linewidth]{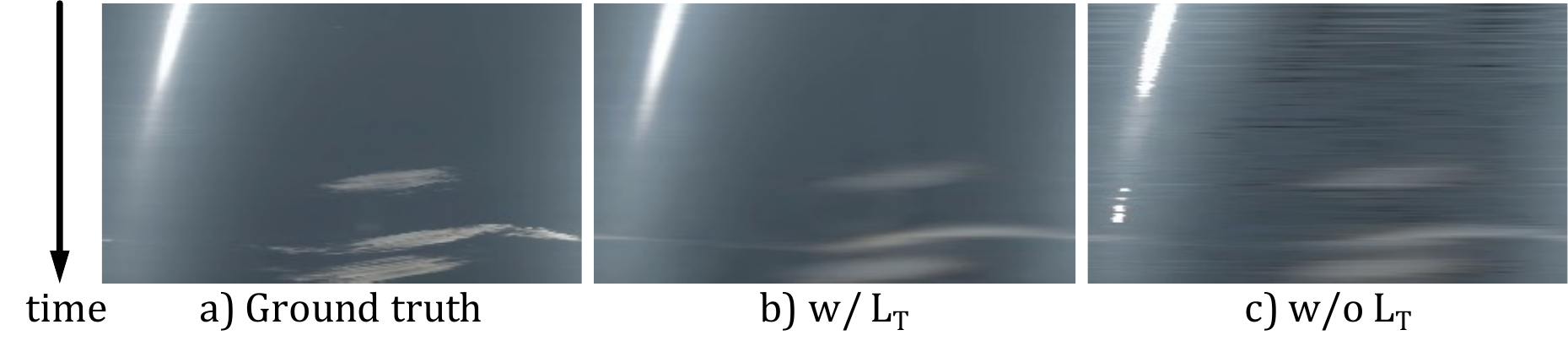}
    \caption{The comparsion between with and without temporal consistency loss $L_T$ using 15 ASGs. The selected row from the HDRI sequence are stacked along the time axis to visualize temporal consistency.}
    \label{fig:dynamic_temporal_stability}
\end{figure}

To validate our composite loss design, we first compare different loss functions for single HDRI fitting, as shown in Fig.~\ref{fig:dynamic_lossfunc}. The experiments demonstrate that L1 loss performs best for capturing both low- and high-frequency lighting signals in the HDRI, as evidenced by the diffuse and specular shading effects in the renderings. In contrast, using L2 loss, as in \cite{tsai2003_sg} or \cite{xu_asg_2013}, even with HDRI intensity preprocessing through functions like log or sRGB, does not yield good results, likely due to the extreme dynamic range of the HDRI. Our diffuse loss $L_D$ leads to accurate recovery of low-frequency lighting, as it is responsible for preserving the global energy of the HDRI. Finally, we demonstrate the necessity of the temporal consistency loss $L_T$ to maintain temporal stability, as shown in Fig.~\ref{fig:dynamic_temporal_stability}. This avoids strong flickering, as we show in our supplemental video.

\section{Conclusion And Future Work}
We propose an optimization method for temporally consistent dynamic HDRI compression preserving all frequencies using ASGs. However, it struggles with fine details in mirror reflections and cannot accurately render sharp-angled lights (e.g. rectangular lights) due to the mixing of spherical Gaussians with higher components.

\bibliographystyle{ACM-Reference-Format}
\bibliography{main}

\end{document}